\documentclass[conference]{IEEEtran}

\usepackage{cite}
\usepackage{amsmath,amssymb,amsfonts}
\usepackage{algorithmic}
\usepackage{graphicx}
\usepackage{textcomp}
\usepackage{xcolor}
\usepackage{booktabs}
\usepackage{multirow}
\usepackage{array}
\usepackage{rotating}
\usepackage{adjustbox}
\usepackage{makecell}
\usepackage{url}
\usepackage[caption=false,font=footnotesize]{subfig}
\usepackage{hyperref}

\newcommand{\best}[1]{\textbf{#1}}

\def\BibTeX{{\rm B\kern-.05em{\sc i\kern-.025em b}\kern-.08em
    T\kern-.1667em\lower.7ex\hbox{E}\kern-.125emX}}

\begin{document}

\title{Beyond Modality Limitations: A Unified MLLM Approach to Automated Speaking Assessment with Effective Curriculum Learning}

\author{\IEEEauthorblockN{Yu-Hsuan Fang, Tien-Hong Lo, Yao-Ting Sung, Berlin Chen}
\IEEEauthorblockA{National Taiwan Normal University\\
\{andyfang, teinhonglo, sungtc, berlin\}@ntnu.edu.tw}}

\maketitle

\begin{abstract}
Traditional Automated Speaking Assessment (ASA) systems exhibit inherent modality limitations: text-based approaches lack acoustic information while audio-based methods miss semantic context. Multimodal Large Language Models (MLLM) offer unprecedented opportunities for comprehensive ASA by simultaneously processing audio and text within unified frameworks. This paper presents a very first systematic study of MLLM for comprehensive ASA, demonstrating the superior performance of MLLM across the aspects of content and language use . However,  assessment on the delivery aspect reveals unique challenges, which is deemed to require specialized training strategies. We thus propose Speech-First Multimodal Training (SFMT), leveraging a curriculum learning principle to establish more robust modeling foundations of speech before cross-modal synergetic fusion. A series of experiments on a benchmark dataset show MLLM-based systems can elevate the holistic assessment performance from a PCC value of 0.783 to 0.846. In particular, SFMT excels in the evaluation of the delivery aspect, achieving an absolute accuracy improvement of 4\% over conventional training approaches, which also paves a new avenue for ASA.
\end{abstract}

\begin{IEEEkeywords}
Multimodal large language model (MLLM), automated speaking assessment (ASA), multimodal training, L2 proficiency, cross-modal learning
\end{IEEEkeywords}

\section{Introduction}

Recent advances in Multimodal Large Language Models (MLLM) have ushered in an unprecedented era of technological transformation, fundamentally reshaping the paradigm of human-machine interaction by jointly integrating information across multiple modalities~\cite{Tang2024SALMONN,Qwen2Audio,Microsoft2025Phi4MiniTR,OmniR1}. Pioneering efforts such as GPT-4o~\cite{openai2024gpt4technicalreport} have demonstrated remarkable capabilities in seamlessly handling text, audio, and visual inputs within an unified framework. Particularly noteworthy is the emergence of open-source MLLM such as Phi-4-multimodal~\cite{Microsoft2025Phi4MiniTR} that has demonstrated superior performance over traditional unimodal approaches after model fine-tuning on domain-specific data~\cite{Devlin2019BERT,Baevski20-Wav2Vec2,WangEQM21,BannoM_SLT2022,MLLM2025feedback_lak} for used in specialized language assessment tasks. Such excellent multimodal capabilities also open new avenues for addressing complex real-world applications previously beyond the reach of conventional approaches.

Within the domain of Computer-Assisted Language Learning (CALL), Automated Speaking Assessment (ASA) represents one of the most challenging and multifaceted tasks~\cite{Lo2024AnEA,deJong2023AssessingSL}. The complexity of evaluating L2 (second-language) speaking proficiency stems from the need to assess multiple aspects of speaking proficiency simultaneously, including delivery (e.g., pronunciation accuracy, fluency, prosodic features), content appropriateness (e.g., topic relevance and coherence), and language use (e.g., vocabulary richness and grammatical correctness)~\cite{Lo2024AnEA,ParkU23}. These evaluation criteria encompass both quantifiable linguistic elements and subtle acoustic characteristics such as stress patterns, intonation contours, and speech rhythm~\cite{Banno2023Assessment,kim22k_interspeech}. The multidimensional nature of speaking assessment, combined with the variability inherent in L2 speech production, establishes ASA systems as indispensable components in modern language learning environments, providing objective, consistent, and scalable evaluation capabilities that complement human assessment~\cite{deJong2023AssessingSL}.

However, traditional ASA approaches suffer from fundamental modality-specific limitations that constrain their effectiveness. Text-based classifiers, exemplified by BERT-based systems~\cite{Devlin2019BERT,WangEQM21}, excel in semantic comprehension and contextual understanding but remain critically dependent on ASR transcription quality and inherently lack access to acoustic features essential for delivery and prosodic evaluation. Conversely, audio-based approaches utilizing self-supervised learning models like wav2vec 2.0~\cite{Baevski20-Wav2Vec2,BannoM_SLT2022} directly process speech signals to capture rich acoustic information for delivery assessment, yet sacrifice semantic context and linguistic content analysis crucial for evaluating language use sophistication and grammatical accuracy. While previous research has explored fusion strategies combining both modalities~\cite{ParkU23}, these approaches typically fuse the outputs of separate unimodal systems, rather than achieving the genuine cross-modal information synchronization found in unified architectures. This fundamental limitation motivates our investigation into whether MLLM can transcend traditional modality boundaries and achieve more effective multimodal integration for comprehensive ASA.

This paper presents a very first systematic study of MLLM for comprehensive ASA, investigating three critical questions: 1) Can multimodal large language models effectively resolve the information fusion challenges encountered in traditional ASA systems, and what performance levels can be achieved? 2) Despite MLLM advances, does the audio modality remain irreplaceable for delivery assessment tasks? 3) Do there exist simple yet cost-effective training strategies that can significantly enhance ASA performance across different aspects of speaking proficiency evaluations? To this end, we design thorough experiments using the TEEMI dataset and propose Speech-First Multimodal Training (SFMT), a curriculum learning approach~\cite{Bengio2009CurriculumL} that progressively transitions from speech foundations to cross-modal integration, achieving an absolute improvement of 4\% in terms of the assessment accuracy for the delivery aspect.

\begin{figure}[!ht]
\centering
\includegraphics[width=\columnwidth]{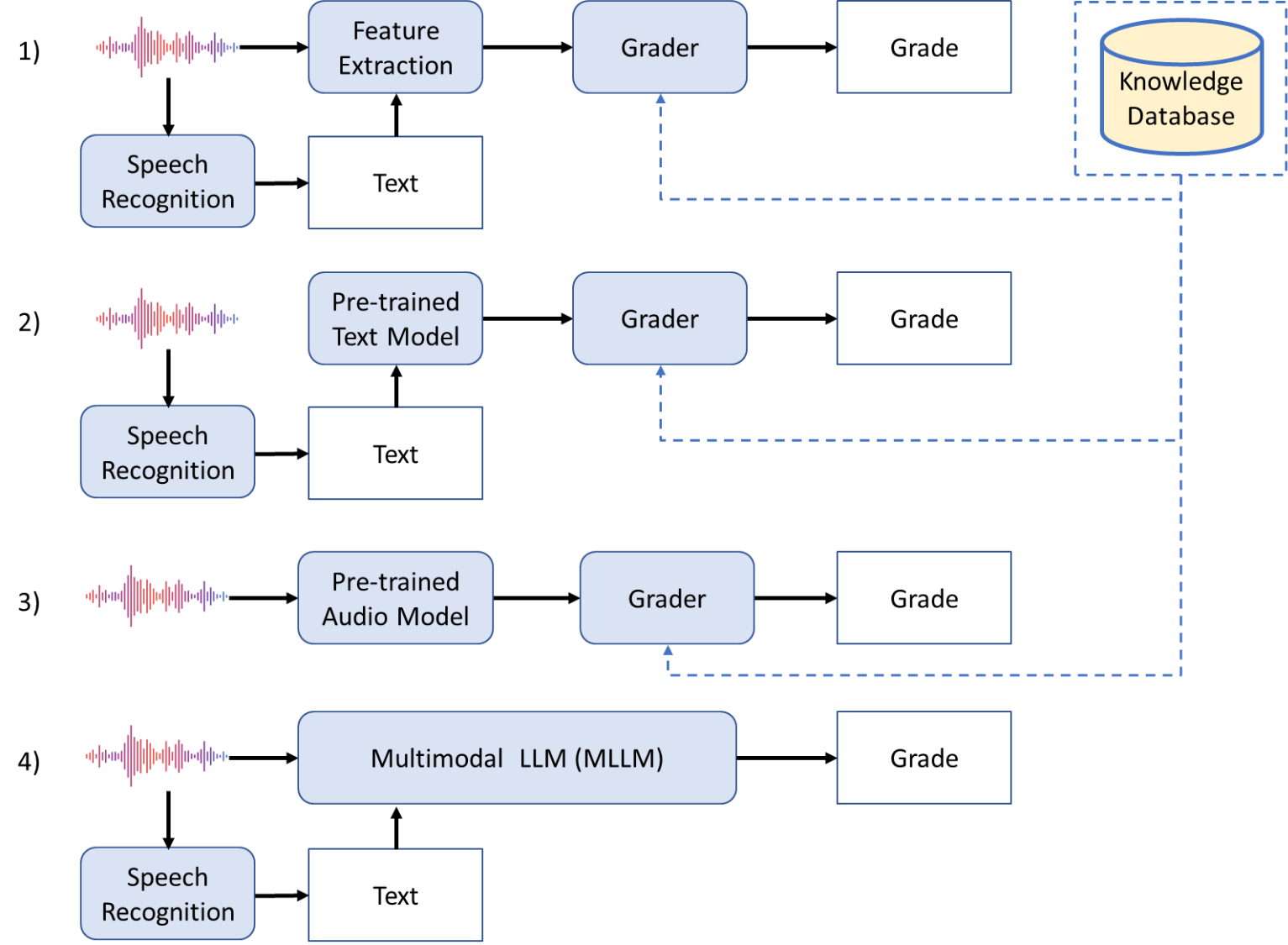}
\caption{ASA systems have evolved from handcrafted feature engineering through self-supervised learning approaches to unified multimodal frameworks capable of comprehensive assessment and feedback generation (adapted from~\cite{Banno2023Assessment}).}
\label{fig:grader_evolution}
\end{figure}

\section{Related Work}

\subsection{Evolution of Automated Speaking Assessment Systems}
Automated speaking assessment (ASA) has evolved through three distinct paradigms, each marking fundamental advances in the automation of evaluations on speaking proficiency for L2 learners. Figure~\ref{fig:grader_evolution} illustrates this progression from handcrafted feature-based systems, through self-supervised models, to unified multimodal frameworks.

\subsubsection{Handcrafted Feature-based Systems}
Early ASA systems typically rely on explicit feature engineering pipelines (Figure~\ref{fig:grader_evolution} (1)), extracting handcrafted acoustic features (spectral, prosodic, temporal) from speech and linguistic features from ASR transcripts~\cite{Davis80-COP}. Traditional machine learning algorithms process these features for proficiency prediction, with Educational Testing Service (ETS) pioneered foundational approaches via extensive feature engineering research~\cite{Loukina2015Feature,Xi2008Automated,Xie2012Exploring}. More recently, Wu et al.~\cite{AesWu2022_rocling} showed that expert-defined knowledge clues (delivery/language use criteria) significantly enhanced assessment performance. Despite interpretability, these systems have limited generalization and require substantial domain expertise.

\subsubsection{Self-Supervised Learning Paradigm}
Self-supervised learning tackles ASA via either text-based or audio-based pre-trained models (Figure~\ref{fig:grader_evolution} (2) and (3)).

\textbf{Text-based Models:} BERT-based models enables sophisticated semantic evaluation (grammar, language use, content) from ASR transcripts~\cite{Devlin2019BERT,WangEQM21}, but are limited by ASR quality and lack acoustic information for assessing the aspect of delivery.

\textbf{Audio-based Models:} Self-supervised speech models like wav2vec 2.0 process raw speech to capture acoustic patterns~\cite{Baevski20-Wav2Vec2,BannoM_SLT2022}. Lo et al.~\cite{Lo2024AnEA} found wav2vec 2.0 inherently encodes syntactic information, revealing the potential of cross-modal feature extraction. Yet, they lack semantic context for comprehensive evaluation.

Both approaches have achieved some success on various ASA tasks, but remain limited by modality constraints. To get around this limitation, prior fusion strategies typically operated at the model level, which would fail to achieve genuine cross-modal synchronization~\cite{ParkU23}.

\subsubsection{Multimodal Large Language Models}
Contemporary MLLM mark a paradigm shift to unified multimodal processing (Figure~\ref{fig:grader_evolution}(4)). Models like Qwen-Audio~\cite{Qwen2Audio}, SALMONN~\cite{Tang2024SALMONN}, and Phi-4-multimodal~\cite{Microsoft2025Phi4MiniTR} simultaneously process speech and text in single frameworks, enabling true multimodal integration via cross-modal attention.

MLLM transcend traditional assessment limitations by providing comprehensive educational feedback beyond scores. Nevertheless, how to design optimal training strategies for multimodal integration, particularly for the evaluation on the aspect of delivery that requires fine-grained acoustic analysis, remains largely underexplored.

\subsection{Curriculum Learning for Multimodal Training}
Curriculum learning posits that structured progression from simple to complex tasks enhances model performance~\cite{Bengio2009CurriculumL}. Recent multimodal speech applications, such as WavLLM~\cite{Chen2024WavLLM} and SALMONN~\cite{Tang2024SALMONN}, have also confirmed the effectiveness of progressive training in speech-text joint modeling. Furthermore, Zhang et al.~\cite{Zhang2024OversamplingAA} applied curriculum learning to speaking assessment via strategic data ordering, showing improvements in limited-data scenarios. However, existing approaches focus on \textit{data-level} curriculum (ordering samples by difficulty), rather than addressing fundamental challenges in multimodal integration.

Our research extends the notion of curriculum learning to \textit{modality-level} progression, investigating the relative importance of acoustic versus textual information for MLLM-based ASA tasks. We propose SFMT, a simple-to-complex learning approach that first establishes robust acoustic foundations before processing cross-modal integration. This modality-level curriculum approach specifically addresses optimizing MLLM performance for fine-grained assessment tasks where acoustic and semantic information must be effectively integrated, while preserving discriminative capabilities essential for accurate proficiency evaluation.

\section{Methodology}

\subsection{Multimodal Large Language Model Architecture for ASA}

\begin{figure}[!t]
    \centering
    \includegraphics[width=0.9\columnwidth]{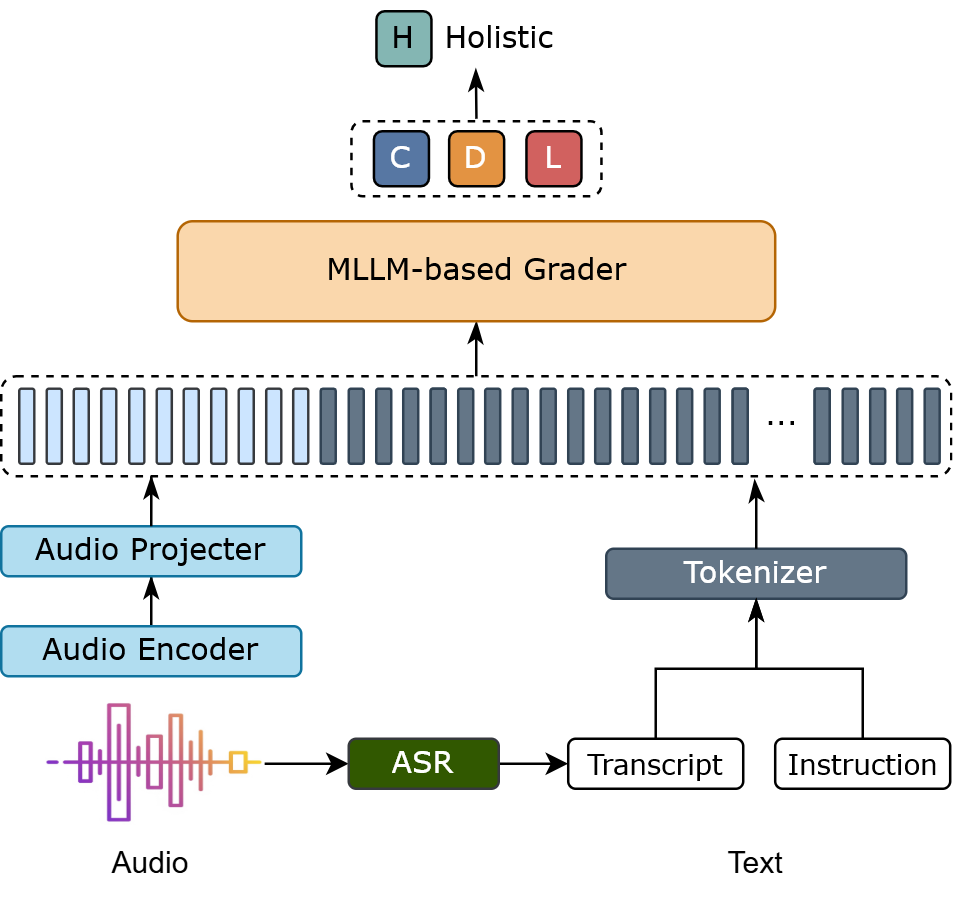}
    \caption{The proposed MLLM architecture processes both audio and text inputs through specialized pathways to generate multi-aspect proficiency scores across content, delivery, language use, and holistic assessment aspects.}
    \label{fig:mllm_architecture}
\end{figure}

We leverage Phi-4-multimodal~\cite{Microsoft2025Phi4MiniTR} for comprehensive automated speaking assessment. This model employs a mixture-of-LoRAs architecture enabling efficient multimodal fine-tuning while preserving base language capabilities. As illustrated in Figure~\ref{fig:mllm_architecture}, the system processes both raw audio and ASR-generated transcripts through modality-specific pathways before integration, comprising: (1) a 3.8B parameter decoder-only Transformer as the reasoning backbone, (2) an audio processing pipeline with 460M-parameter encoder using conformer blocks and audio projector for shared embedding space mapping, and (3) a modality-specific audio adapter (LoRA$_{\text{audio}}$, 460M parameters) enabling learning of targeted acoustic traits without language capability interference.

For comprehensive assessment on the spoken responses of the TEEMI dataset, we train three specialized models targeting aspects of Content (C), Delivery (D), and Language Use (L), respectively. Each model receives aspect-specific instructions during training, allowing focused optimization. The Holistic (H) score integrates assessment results gathered from all three aspects, providing an overall proficiency indicator aligned with CEFR standards.

\subsection{Speech-First Multimodal Training (SFMT) Strategy}

Standard multimodal training approach to ASA encounters a fundamental challenge: modality imbalance. Models of these approaches tend to exhibit systematic preference for textual features due to their structured representations and computational efficiency, consequently underutilizing acoustic information critical for delivery assessment~\cite{modality_imbalance2023cvpr}. This imbalance impairs the model's capacity to learn fine-grained acoustic patterns—including pronunciation accuracy, fluency variations, and prosodic characteristics—that text representations inherently cannot encode.

Our empirical investigation through systematic ablation studies (Section~\ref{subsec:modality_analysis}) reveals a counterintuitive finding: the audio modality demonstrates superior learning efficiency for MLLM-based graders compared to text, particularly for the assessment on the delivery aspect. This observation of audio's stronger initial performance and faster convergence under identical training conditions motivates our speech-first strategy.

This empirical superiority of acoustic learning stems from three fundamental factors:

\textbf{(1) Information Completeness}: Raw audio signals preserve the complete spectrum of speech information—from phonetic details to prosodic contours—providing MLLM with unfiltered access to all acoustic evidence necessary for proficiency assessment. In contrast, ASR-transcribed text represents a lossy transformation that discards paralinguistic features critical for delivery evaluation.

\textbf{(2) Direct Signal Access}: Audio inputs bypass the error propagation inherent in text-based approaches, offering direct access to ground-truth acoustic patterns. This eliminates the cascading effects of ASR transcription errors and systematic biases from ASR systems trained predominantly on native speech.

\textbf{(3) Preferential Learning Patterns}: When exposed to both modalities simultaneously, models demonstrate preferential optimization toward text-based features as computationally efficient pathways~\cite{yu2024emnlp}, particularly for content and language use assessment. This preference inhibits the development of acoustic discrimination capabilities, as models converge on solutions that underutilize acoustic information.

\begin{figure*}[!t]
    \centering
    \includegraphics[width=0.85\textwidth]{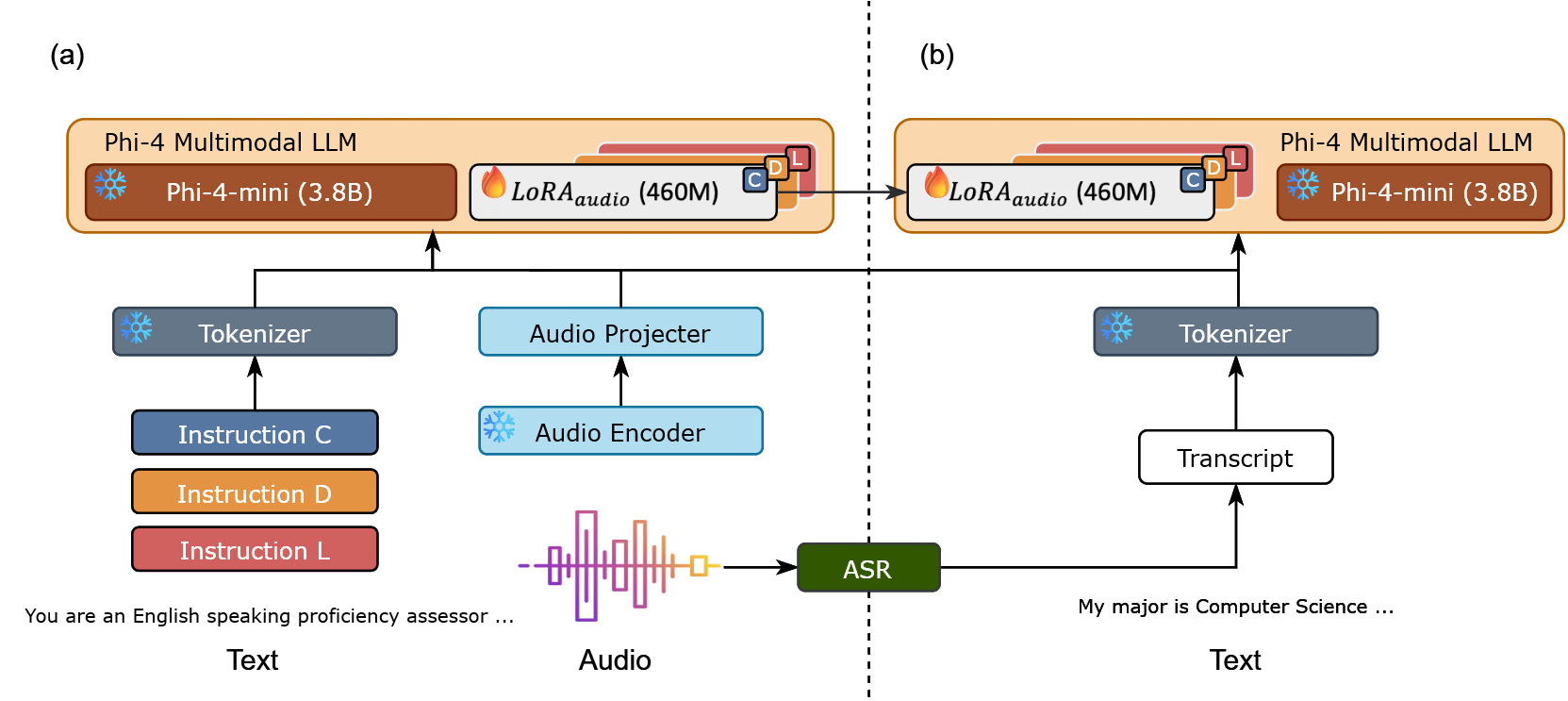}
    \caption{SFMT employs a two-stage curriculum learning approach that first establishes acoustic foundations through audio-only training before introducing cross-modal integration with textual information.}
    \label{fig:afmt_architecture}
\end{figure*}

Building upon these insights, we propose Speech-First Multimodal Training (SFMT), a two-stage curriculum learning strategy that exploits the discovered learning hierarchy. By establishing robust acoustic feature extraction capabilities before introducing textual information, SFMT ensures that models develop strong delivery assessment abilities that persist through subsequent multimodal integration(Figure~\ref{fig:afmt_architecture}):

\textbf{Stage 1 - Acoustic Foundation (Fig.~\ref{fig:afmt_architecture}(a)):}
Given training data $\mathcal{D}_{\text{audio}} = \{(\mathbf{a}_i, I_i, y_i)\}_{i=1}^N$ where $\mathbf{a}_i$ is audio input vector, $I_i \in \{I_C, I_D, I_L\}$ is aspect-specific instruction, and $y_i$ is the target score, we optimize:
\begin{equation}
\theta_{\text{LoRA}}^{1} = \arg\min_{\theta_{\text{LoRA}}} \sum_{(\mathbf{a},I,y) \in \mathcal{D}_{\text{audio}}} \mathcal{L}(f_{\text{Phi-4}}(\mathbf{a}, I; \theta_{\text{LoRA}}), y),
\label{eq:stage1_afmt}
\end{equation}
where $f_{\text{Phi-4}}$ denotes the MLLM and $\mathcal{L}$ is the loss function. Only the LoRA audio adapter parameters $\theta_{\text{LoRA}}$ are updated.

\textbf{Stage 2 - Cross-Modal Integration (Fig.~\ref{fig:afmt_architecture}(b)):}
Using multimodal data $\mathcal{D}_{\text{multi}} = \{(\mathbf{a}_i, \mathbf{t}_i, I_i, y_i)\}_{i=1}^N$ with additional transcript vector $\mathbf{t}_i$, we continue optimization from Stage 1:
\begin{equation}
\theta_{\text{LoRA}}^{2} = \arg\min_{\theta_{\text{LoRA}}^{1}} \sum_{(\mathbf{a},\mathbf{t},I,y) \in \mathcal{D}_{\text{multi}}} \mathcal{L}(f_{\text{Phi-4}}(\mathbf{a}, \mathbf{t}, I; \theta_{\text{LoRA}}^{1}), y),
\label{eq:stage2_afmt}
\end{equation}
where $\theta_{\text{LoRA}}$ is the pre-trained adapter. This progression ensures robust acoustic specialization before multimodal integration, particularly enhancing the performance of the assessment on the delivery aspect.

\section{Experiments}

\subsection{Datasets}
We evaluate our proposed models on two distinct datasets: the proprietary TEEMI corpus and the publicly available  the Speak \& Improve Corpus.

\subsubsection{TEEMI Corpus}

The TEEMI corpus (Test for English-Medium Instruction)~\cite{CorpusChen2024_teemi_interspeech} is a comprehensive L2 proficiency dataset designed for EMI research in higher education contexts. The corpus features spontaneous English speech from undergraduate and graduate L2 learners, with each response evaluated across four aspects—holistic, content, language use, and delivery—using an eight-level CEFR-aligned scale (Pre-A1 to B2).  TEEMI is equipped with triple-rater annotation with majority voting to ensure scoring reliability.

The speaking assessment of TEEMI includes three task formats: general listen and answer (A), situational question and answer (B), and thematic question and answer (C). In this paper, we focus on a subset consisting of tasks A01, A02, yielding a total of 8,214 responses. Model training and validation are performed solely on A01, which contains 6,152 responses from 1,231 speakers. The A02 task is held out to evaluate the model’s ability to generalize to previously unseen prompts. The detailed CEFR-level distributions for the A01 and A02 tasks utilized in this study are illustrated in Table~\ref{tab:teemi_dataset_selected}.

\begin{table}[!ht]
\centering
\caption{Statistical information for selected CEFR proficiency levels (A01, A02) in the TEEMI dataset.}
\label{tab:teemi_dataset_selected}
\resizebox{\columnwidth}{!}{%
\begin{tabular}{lcccclllll}
\hline
\textbf{Task} & \textbf{Usage} & \textbf{Pre-A} & \textbf{A1} & \multicolumn{1}{l}{\textbf{A1+}} & \textbf{A2} & \textbf{A2+} & \textbf{B1} & \textbf{B1+} & \textbf{B2} \\ \hline
\multirow{3}{*}{A01} & Train  & 34 & 61  & 76  & 156 & 150 & 169 & 79  & 65  \\
                     & Valid  & 8  & 16  & 19  & 38  & 39  & 43  & 23  & 12  \\
                     & Test   & 11 & 20  & 23  & 49  & 50  & 48  & 32  & 15  \\ \hline
A02                  & Unseen & 9  & 7   & 12  & 19  & 12  & 26  & 23  & 15  \\ \hline
Total                & \textbf{-}      & 62 & 104 & 130 & 262 & 251 & 286 & 157 & 107 \\ \hline
\end{tabular}%
}
\end{table}

\subsubsection{SLaTE 2025 Speak \& Improve Corpus}

We utilize the Speak \& Improve Corpus 2025~\cite{Knill2024SpeakI}, containing 315 hours of L2 English speech with CEFR proficiency levels from A2 to C1+. The corpus includes four task types: Interview, Opinion, Presentation, and Communication Activity, equipped with holistic scores averaged across different aspects. We follow official data splits to construct the corresponding training, development, and test sets.

\begin{table*}[!ht]
\centering
\caption{Model performance on the TEEMI test set.}
\label{tab:main_results_8scale}
\adjustbox{width=\textwidth}{
\begin{tabular}{@{}l rrr rrr rrr rrr@{}}
\toprule
\multirow{2}{*}{\textbf{Models}} & \multicolumn{3}{c}{\textbf{Content (C)}} & \multicolumn{3}{c}{\textbf{Delivery (D)}} & \multicolumn{3}{c}{\textbf{Language Use (L)}} & \multicolumn{3}{c}{\textbf{Holistic (H)}} \\
\cmidrule(lr){2-4} \cmidrule(lr){5-7} \cmidrule(lr){8-10} \cmidrule(lr){11-13}
& \textbf{PCC↑} & \textbf{ABS↑} & \textbf{ADJ↑} & \textbf{PCC↑} & \textbf{ABS↑} & \textbf{ADJ↑} & \textbf{PCC↑} & \textbf{ABS↑} & \textbf{ADJ↑} & \textbf{PCC↑} & \textbf{ABS↑} & \textbf{ADJ↑} \\
\midrule
\multicolumn{13}{@{}l}{\textbf{Baseline Models}} \\
W2V~\cite{Baevski20-Wav2Vec2} & 0.755 & 35.08 & 81.85 & 0.768 & 39.92 & 83.06 & 0.740 & 36.29 & 79.03 & 0.771 & 34.67 & 83.87 \\
BERT~\cite{Devlin2019BERT} & 0.774 & 33.47 & 84.68 & 0.794 & 38.31 & 84.68 & 0.759 & 36.29 & 80.24 & 0.781 & 35.48 & 82.66 \\
W2V-BERT~\cite{ParkU23} & 0.735 & 35.08 & 81.45 & 0.794 & 38.71 & 87.10 & 0.798 & 41.13 & 82.66 & 0.771 & 38.71 & 84.68 \\
W2V-PT~\cite{Lo2024AnEA} & 0.733 & 30.65 & 79.84 & 0.796 & 39.11 & 83.06 & 0.779 & \best{42.74} & 81.45 & 0.785 & 34.68 & 83.07 \\
BERT-PT~\cite{Lo2024AnEA} & 0.756 & 29.44 & 79.84 & 0.783 & 40.73 & 83.06 & 0.788 & 35.08 & 81.85 & 0.777 & 33.87 & 81.85 \\
Multi-Aspect~\cite{Peng2024EnhancingAS} & 0.760 & 37.10 & 80.24 & 0.810 & 41.94 & 85.48 & 0.785 & 39.92 & 81.45 & 0.783 & 38.31 & 84.27 \\
\midrule
\multicolumn{13}{@{}l}{\textbf{Our Approach}} \\
Phi-4 & \best{0.826} & \best{41.93} & 87.90 & 0.831 & 42.34 & \best{89.11} & \best{0.840} & 41.53 & \best{89.52} & \best{0.846} & \best{42.34} & 90.32 \\
Phi-4 (SFMT) & 0.821 & 39.11 & \best{88.31} & \best{0.848} & \best{46.77} & \best{89.11} & 0.835 & 40.73 & 88.31 & 0.838 & 41.13 & \best{90.73} \\
\bottomrule
\end{tabular}%
}
\end{table*}

\begin{figure*}[!t]
\centering
\subfloat[Content (Phi-4)\label{fig:cm_content_phi4}]{\includegraphics[width=0.225\textwidth]{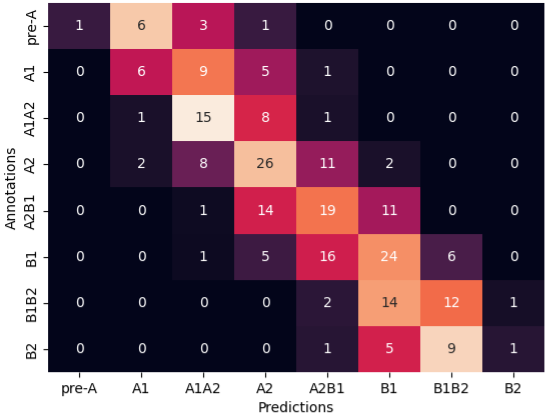}}
\hfill
\subfloat[Delivery (Phi-4)\label{fig:cm_pronunciation_phi4}]{\includegraphics[width=0.225\textwidth]{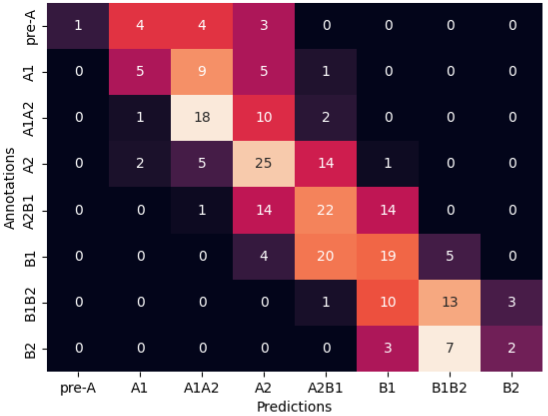}}
\hfill
\subfloat[Language use (Phi-4)\label{fig:cm_vocabulary_phi4}]{\includegraphics[width=0.225\textwidth]{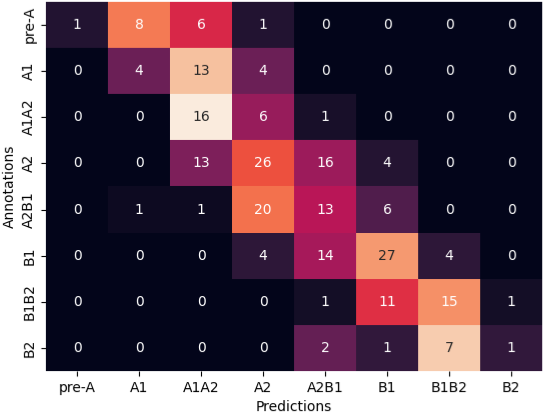}}
\hfill
\subfloat[Holistic (Phi-4)\label{fig:cm_holistic_phi4}]{\includegraphics[width=0.225\textwidth]{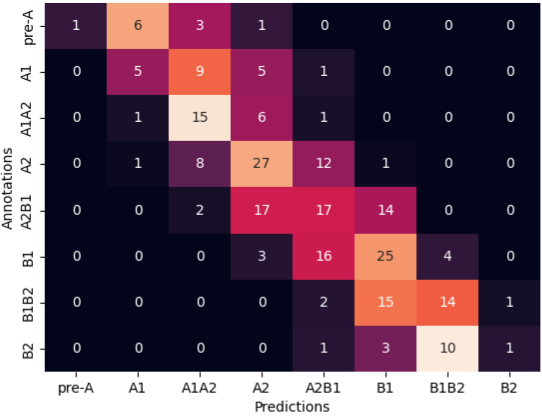}}

\vspace{0.01cm} 

\subfloat[Content (SFMT)\label{fig:cm_content_afmt}]{\includegraphics[width=0.225\textwidth]{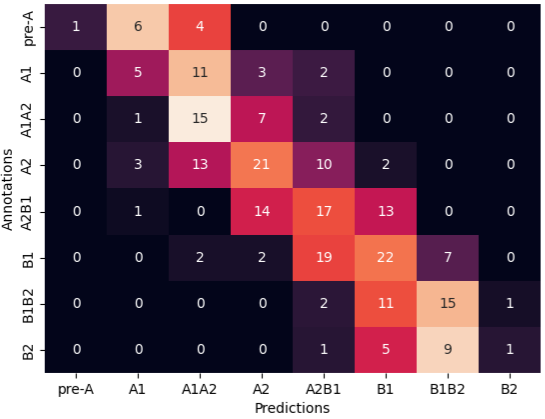}}
\hfill
\subfloat[Delivery (SFMT)\label{fig:cm_pronunciation_afmt}]{\includegraphics[width=0.225\textwidth]{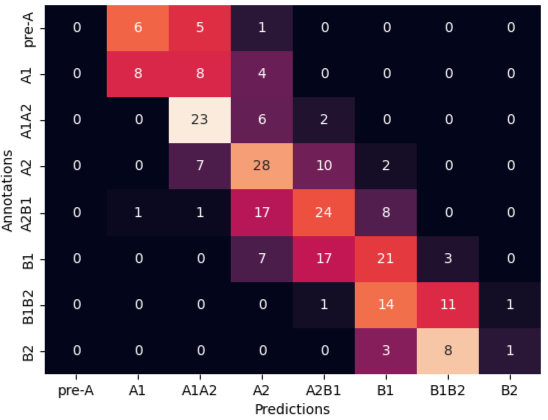}}
\hfill
\subfloat[Language use (SFMT)\label{fig:cm_vocabulary_afmt}]{\includegraphics[width=0.225\textwidth]{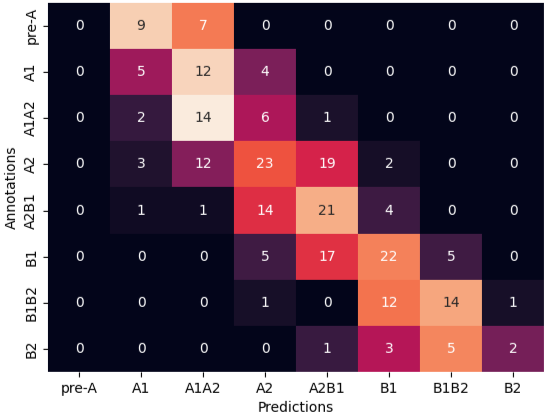}}
\hfill
\subfloat[Holistic (SFMT)\label{fig:cm_holistic_afmt}]{\includegraphics[width=0.225\textwidth]{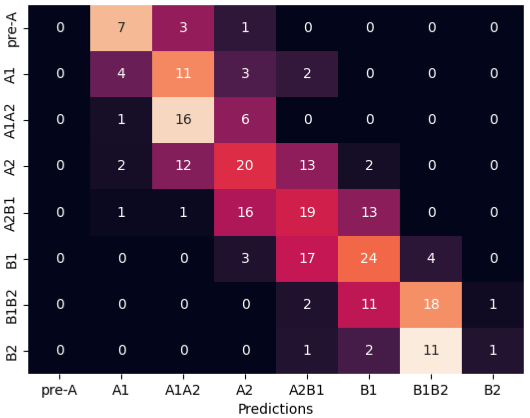}}

\caption{Confusion matrices comparing standard Phi-4 and SFMT performance on CEFR scale, demonstrating enhanced diagonal concentration particularly for delivery assessment.}
\label{fig:confusion_matrices}
\end{figure*}

\subsection{Implementation Details}

Model configurations were initialized using the \texttt{Phi-4-multimodal-instruct}\footnote{\url{https://huggingface.co/microsoft/Phi-4-multimodal-instruct}} with LoRA adaptation~\cite{hu2022lora} (rank=320) applied to the audio encoder. Training employed the AdamW optimizer (lr=4e-5) for 3 epochs with batch size 32 (gradient accumulation steps: 16) and bfloat16 mixed precision on a single NVIDIA RTX 3090. Flash attention~\cite{dao2024flashattention} was utilized for memory efficiency.

For speech recognition, we compare Whisper large v2 (14.75\% WER) against the integrated ASR module of Phi-4 (18.25\% WER) on the TEEMI corpus. Output generation is constrained to 10 tokens to prevent hallucination. SFMT training follows the prescribed two-stage curriculum: Stage 1 processes audio-only inputs with null text placeholders, while Stage 2 incorporates full multimodal inputs. Aspect-specific prompts guide targeted assessment during inference.

Model performance is evaluated using Pearson Correlation Coefficient (PCC) for prediction consistency, Absolute Accuracy for exact CEFR-level classification, Adjacent Accuracy for predictions within ±0.5 levels, and Macro Accuracy for balanced cross-level performance measurement accounting for dataset class imbalance. Additionally, for the evaluation of regression-based scoring tasks, Root Mean Squared Error (RMSE) is utilized to assess the average magnitude of the error between predicted and actual continuous scores.

To facilitate reproducibility and promote community advancement in multimodal ASA research, we will make all source code and fine-tuning implementations publicly available upon publication\footnote{\url{https://github.com/ntnuYuhsuan/asa-grader.git}}.

\section{Results}

\subsection{Overall MLLM Performance}

Table~\ref{tab:main_results_8scale} demonstrates substantial MLLM (viz. Phi-4) superiority over current state-of-the-art models across all aspects of assessment. Standard Phi-4 achieves PCC scores consistently above 0.82, representing significant improvements over the compared models which all have PCC results below 0.80. This seems to validate MLLM multimodal integration capabilities for comprehensive ASA.

The confusion matrices in Figure~\ref{fig:confusion_matrices} provide visual confirmation of enhanced classification precision when performing ASA with the MLLM-based models; MLLM-based models exhibits superior diagonal concentration compared to traditional ones, suggesting MLLM as an all-around workhorse capable of transcending inherent modality limitations.

\subsection{Modality Analysis and SFMT Effectiveness}
\label{subsec:modality_analysis}

The ablation studies, with Pearson Correlation Coefficient (PCC) and Macro Accuracy (Macro Acc) as key performance indicators (Table~\ref{tab:ablation}), reveal fundamental insights into modality contributions for MLLM-based ASA graders and validates the efficacy of our SFMT strategy.

\begin{table*}[!htbp]
\centering
\caption{Ablation study comparing modality contributions to MLLM-based ASA performance.}
\label{tab:ablation}
\begin{tabular}{@{}lcccccccccc@{}}
\toprule
\multirow{2}{*}{\textbf{Training Configuration}} & \multirow{2}{*}{\textbf{Audio}} & \multirow{2}{*}{\textbf{Text}} & \multicolumn{2}{c}{\textbf{Content (C)}} & \multicolumn{2}{c}{\textbf{Delivery (D)}} & \multicolumn{2}{c}{\textbf{Language Use (L)}} & \multicolumn{2}{c}{\textbf{Holistic (H)}} \\
\cmidrule(lr){4-5} \cmidrule(lr){6-7} \cmidrule(lr){8-9} \cmidrule(lr){10-11}
& & & \textbf{PCC↑} & \textbf{Macro Acc↑} & \textbf{PCC↑} & \textbf{Macro Acc↑} & \textbf{PCC↑} & \textbf{Macro Acc↑} & \textbf{PCC↑} & \textbf{Macro Acc↑} \\
\midrule
Phi-4 & \checkmark & \checkmark & \best{0.826} & 82.00 & 0.831 & 82.48 & \best{0.840} & 85.27 & \best{0.841} & 84.27 \\
\midrule
Phi-4 (Text-Only) & × & \checkmark & 0.784 & 74.76 & 0.776 & 75.83 & 0.768 & 72.80 & 0.776 & 73.35 \\
Phi-4 (Audio-Only) & \checkmark & × & 0.811 & 82.33 & 0.835 & 82.94 & 0.830 & \best{86.16} & 0.836 & 86.83 \\
Phi-4 (SFMT) & \checkmark & \checkmark & 0.821 & \best{83.41} & \best{0.848} & \best{84.01} & 0.835 & 83.67 & 0.838 & \best{86.75} \\
\bottomrule
\end{tabular}
\end{table*}

\textbf{Modality Contributions}: Table~\ref{tab:ablation} reports on the  performance levels of MLLM-based models that operate on different modalities and their combination. Audio-only configurations demonstrate strong overall performance, particularly excelling in the assessment on the delivery aspect. In contrast, text-only models exhibit a general decline in performance, with the most significant drop observed in the assessment on the delivery aspect. This underscores the challenges facing text-only models, which is partly due to  their reliance on ASR transcripts alone (achieving 14.75\% WER with Whisper large v2 on TEEMI) and the inherent lack of direct acoustic cues for the assessment on the delivery aspect.

\textbf{SFMT Validation}: The strategic emphasis of SFMT on establishing robust speech processing foundations before introducing textual information yields significant enhancements, particularly in the assessment on the delivery aspect—whose success is most critically dependent on fine-grained acoustic discrimination. This is clearly demonstrated by improvements over the Phi-4 baseline (Table~\ref{tab:ablation}): the assessment on the delivery aspect shows a pronounced PCC advantage (a value of 0.848 for SFMT vs. 0.831 for the Phi-4 baseline). Furthermore, SFMT improves on the Macro Accuracy for this aspect from 82.48\% to 84.01\%. These results validate SFMT as an effective curriculum learning approach, highlighting the benefits of establishing robust acoustic representations prior to cross-modal integration.

\subsection{Generalization to Unseen Tasks}
Evaluation on the unseen tasks of TEEMI (\textit{cf.} Table~\ref{tab:unseen_results}) confirms the robust generalization capablilty of fine-tuned Phi-4 across all aspects. The assessment on the delivery aspect exhibits the strongest transfer performance, indicating effective learning of transferable acoustic features. The results on the content and language use aspects also show strong correlations despite semantic variations in task prompts. This again validates the MLLM's capability to develop generalizable multimodal representations for cross-task ASA applications.

\begin{table}[!ht]
\centering
\caption{Model performance on the unseen TEEMI dataset.}
\label{tab:unseen_results}
\begin{tabular}{@{}l ccc@{}}
\toprule
\textbf{Aspect} & \textbf{PCC↑} & \textbf{ABS Acc↑} & \textbf{ADJ Acc↑} \\
\midrule
Content (C)      & 0.851         & 32.52             & 78.86             \\
Delivery (D) & 0.863         & 44.72             & 86.18             \\
Language Use (L)   & 0.855         & 33.33             & 78.86             \\
Holistic (H)     & 0.846         & 32.52             & 78.86             \\
\bottomrule
\end{tabular}
\end{table}

\subsection{Cross-corpus evaluation}
Cross-corpus evaluation on the Speak \& Improve Corpus (Table~\ref{tab:slate_results}) further confirms the effectiveness of our model across diverse L2 populations and assessment tasks. The SFMT strategy consistently outperforms both the traditional baselines and the standard Phi-4 implementation across all evaluation metrics, demonstrating superior prediction accuracy and correlation with human judgments. This cross-corpus success validates that the proposed model and training regime generalize beyond the specific characteristics of the TEEMI corpus to broader international assessment contexts. The consistent performance improvements across different datasets and learner populations establish the practical applicability of SFMT for real-world ASA deployment scenarios.

\begin{table}[!ht]
\centering
\caption{Performance on the Speak \& Improve Corpus.}
\label{tab:slate_results}
\begin{tabular}{lcccc}
\toprule
\textbf{Method} & \textbf{RMSE↓} & \textbf{PCC↑} & \textbf{Acc±0.5↑} & \textbf{Acc±1.0↑} \\
\midrule
BERT~\cite{Devlin2019BERT} & 0.445 & 0.727 & 76.0 & 96.3 \\
W2V~\cite{Baevski20-Wav2Vec2} & 0.394 & 0.790 & \best{81.3} & \best{99.3} \\
Phi-4 & 0.412 & 0.796 & 74.7 & 98.0 \\
Phi-4 (SFMT) & \best{0.387} & \best{0.800} & 79.7 & 99.2 \\
\bottomrule
\end{tabular}
\end{table}

\section{Conclusion and Future Work}

This paper presents a very first systematic study of MLLM for comprehensive automated speaking assessment (ASA), addressing three fundamental research questions. Our findings demonstrate that MLLM effectively resolve traditional  challenges facing information fusion, achieving superior performance across all assessment aspects compared to uni-modality based models. The ablation studies confirm the irreplaceability of the audio modality for delivery assessment, while the proposed SFMT strategy considerably promotes performance through speech-first curriculum learning, particularly benefiting fine-grained acoustic discrimination. A series of experimental validation on TEEMI and the Speak \& Improve Corpus confirm the robust generalization capability of our model  across diverse L2 populations and assessment contexts. These results also suggest MLLM-based models as the transformative backbone for ASA, enabling more accurate, comprehensive, and generalizable evaluation systems. Future research will explore multi-task learning frameworks for multi-aspect assessment and integrate comprehensive feedback generation into ASA, advancing towards the broader goal of creating intelligent, adaptive language learning environments that can provide personalized, real-time guidance for L2 learners in various contexts of computer-assisted language learning.

\bibliographystyle{IEEEtran}
\bibliography{references}

\end{document}